\documentclass[final]{cvpr}

\usepackage{times}
\usepackage{epsfig}
\usepackage{graphicx}
\usepackage{amsmath}
\usepackage{amssymb}

\usepackage{booktabs}
\usepackage{caption}
\usepackage{subcaption}
\usepackage{multirow}
\usepackage{bm}
\usepackage{amsmath}
\usepackage[dvipsnames]{xcolor}

\DeclareMathOperator*{\argmax}{arg\,max}

\definecolor{myorange}{RGB}{247, 152, 63}
\definecolor{mygreen}{RGB}{0, 153, 77}

\usepackage[pagebackref=true,breaklinks=true,colorlinks,bookmarks=false]{hyperref}
\def\*#1{\mathbf{#1}}

\def\cvprPaperID{6415} 
\def\confYear{CVPR 2021}

\begin{document}

\title{MOS: Towards Scaling Out-of-distribution Detection for Large Semantic Space}

\author{Rui Huang\\
Department of Computer Sciences\\
University of Wisconsin-Madison\\
{\tt\small huangrui@cs.wisc.edu}
\and
Yixuan Li\\
Department of Computer Sciences\\
University of Wisconsin-Madison\\
{\tt\small sharonli@cs.wisc.edu}
}
\maketitle

\begin{abstract}
\vspace{-0.2cm}
Detecting out-of-distribution (OOD) inputs is a central challenge for safely deploying machine learning models in the real world. 
Existing solutions are mainly driven by small datasets, with low resolution and very few class labels (\eg, CIFAR). As a result, OOD detection for large-scale image classification tasks remains largely unexplored. In this paper, we bridge this critical gap by proposing a group-based OOD detection framework, along with a novel OOD scoring function termed \textbf{MOS}. Our key idea is to decompose the large semantic space into smaller groups with similar concepts,  which allows simplifying the decision boundaries between in- vs. out-of-distribution data for effective OOD detection. Our method scales substantially better for high-dimensional class space than previous approaches. We evaluate models trained on ImageNet against four carefully curated OOD datasets, spanning diverse semantics. MOS  establishes state-of-the-art performance, reducing the average FPR95 by 14.33\% while achieving {6x} speedup in inference compared to the previous best method.



\end{abstract}

\begin{figure*}[t]
    \centering
    \vspace{-0.2cm}
    \includegraphics[width=0.92\textwidth]{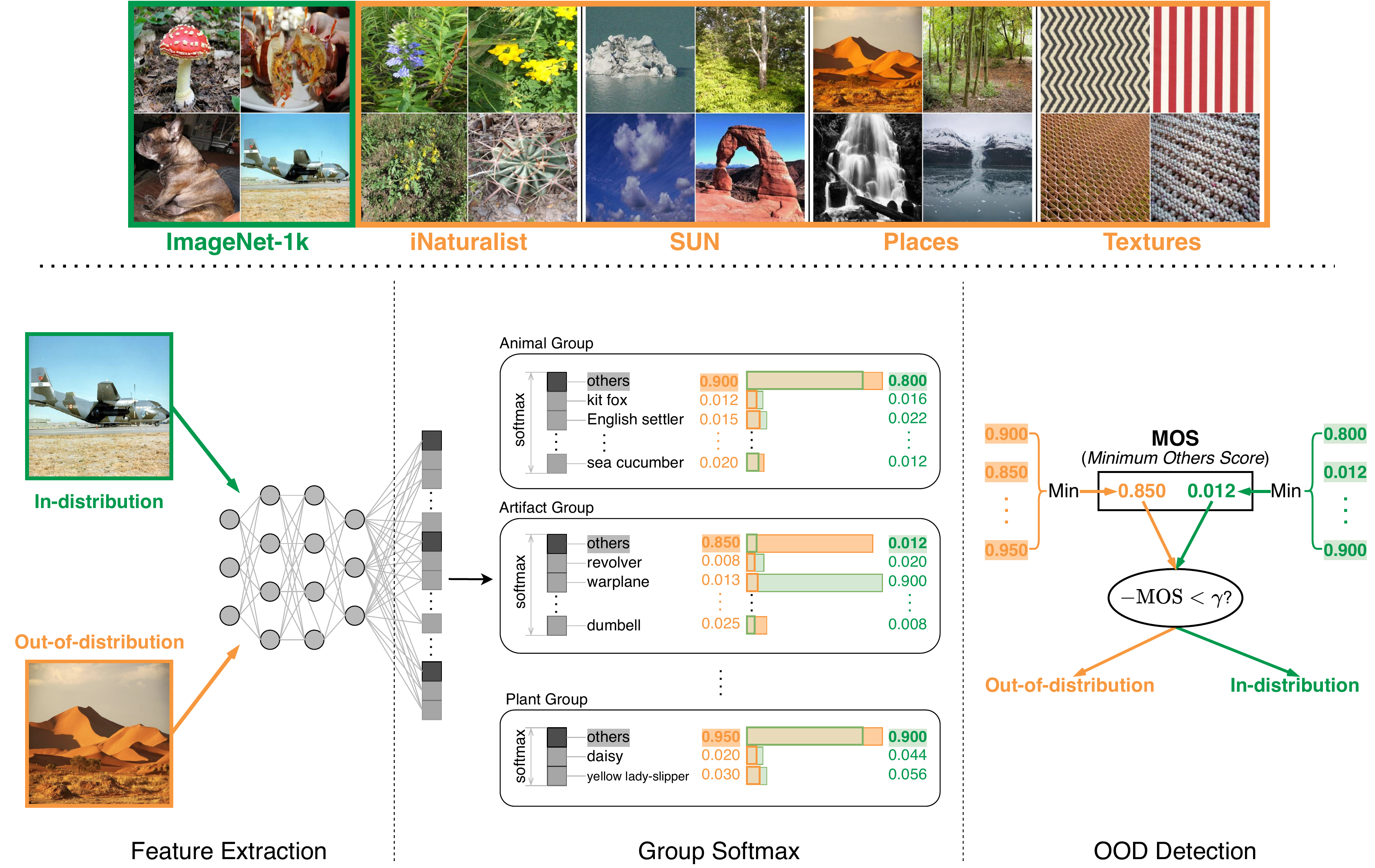}
    \caption{\small \textit{Top}: Examples of in-distribution images sampled from ImageNet-1k (in {\textcolor{mygreen}{green}}) and  OOD images sampled from 4 datasets described in Section~\ref{sec:dataset} (in {\textcolor{myorange}{orange}}). \textit{Bottom}: Overview of the proposed group-based OOD detection framework. The key idea is to decompose the large semantic space into smaller groups, which allows simplifying the decision boundary between in- and out-of-distribution data. A category \texttt{others} is added to each group. An OOD image is mapped to \texttt{others} with high confidence for all groups, whereas an in-distribution image will have a lower score for \texttt{others} in the group it belongs to (\eg, \texttt{artifact} group). The minimum score on category \texttt{others} among all groups, MOS, allows effective differentiation of OOD data.  
    }
    \vspace{-0.45cm}
    \label{fig:main_arch}
\end{figure*}

\vspace{-0.4cm}
\section{Introduction}
Out-of-distribution (OOD) detection has become a central challenge in safely deploying machine learning models in the open world, where the test data may be distributionally different from the training data. A plethora of literature has emerged in addressing the problem of OOD detection~\cite{bevandic2018discriminative,hein2019relu, hendrycks2016baseline,
lakshminarayanan2017simple,
lee2018simple,
liang2018enhancing, mohseni2020self, chen2020robust-new, hsu2020generalized, liu2020energy, lin2021mood}. However, existing solutions are mainly driven by small, low-resolution datasets such as CIFAR~\cite{krizhevsky2009learning} and MNIST~\cite{lecun2010mnist}. Deployed systems like autonomous vehicles often operate on images that have
far greater resolution and perceive environments with far
more categories. As a result, a critical research gap exists in developing and evaluating OOD detection algorithms for large-scale image classification tasks. 

While one may be eager to conclude that solutions for small datasets should transfer to a large-scale setting, we argue that this is far from the truth. The main challenges posed in OOD detection stem from the fact that it is impossible to comprehensively define and anticipate anomalous data in advance, resulting in a large space of uncertainty. As the number of semantic classes increases, the plethora of ways that OOD data may occur increases correspondingly. For example, our analysis reveals that the average false positive rate (at 95\% true positive rate) of a common baseline~\cite{hendrycks2016baseline} would rise from 17.34\% to 76.94\% as the number of classes increases from 50 to 1,000 on ImageNet-1k~\cite{deng2009imagenet}. Very few works have studied OOD detection in the large-scale setting, with limited evaluations and effectiveness~\cite{roady2019outofdistribution,hendrycks2019benchmark}. This begs the following question: \emph{how can we design an OOD detection algorithm that scales effectively for classification with large semantic space}?


Motivated by this, we take an important step to bridge this gap and propose a group-based OOD detection framework that is effective for large-scale image classification. Our key idea is to decompose the large semantic space into smaller groups with similar concepts, which allows simplifying the decision boundary and reducing the uncertainty space between in- vs. out-of-distribution data. Intuitively, for OOD detection, it is simpler to estimate whether an image belongs to one of the coarser-level semantic groups than to estimate whether an image belongs to one of the finer-grained classes. For example, consider a  model tasked with classifying 200 categories of plantations and another 200 categories of marine animals. A \texttt{truck} image can be easily classified as OOD data since it does not resemble either the plantation group or the marine animal group.

Formally, our proposed method leverages group softmax and derives a novel OOD scoring function. Specifically, the group softmax computes probability distributions within each semantic group. A key component is to utilize a category \texttt{others} in each group, which measures the probabilistic score for an image to be OOD with respect to the group. Our proposed  OOD scoring function, \emph{Minimum Others Score} (\textbf{MOS}), exploits the information carried by the \texttt{others} category. As illustrated in Figure~\ref{fig:main_arch}, MOS is higher for OOD inputs as they will be mapped to \texttt{others} with high confidence in all groups, and is lower for in-distribution inputs.


We extensively evaluate our approach on models trained with the ImageNet-1k dataset, leveraging the state-of-the-art pre-trained BiT-S models~\cite{kolesnikov2020big} as backbones. We explore label space of size 10-100 times larger than that of previous works~\cite{hendrycks2016baseline,liang2018enhancing,lee2018simple,liu2020energy,chen2020robust-new,hendrycks2018deep}. Compared to the best baseline~\cite{hendrycks2019benchmark}, our method improves the average performance of OOD detection by 14.33\% (FPR95) over four diverse OOD test datasets. 
More importantly, our method achieves improved OOD detection performance while preserving the classification accuracy on in-distribution datasets. We note that while group-based learning has been used for improving tasks such as long-tail object detection~\cite{li2020overcoming}, our objective and motivation are very different---we are interested in reducing the uncertainty between in- and out-of-distribution data, rather than reducing the confusion among in-distribution data themselves. 
Below we summarize our \textbf{key results and contributions}:
\vspace{-0.1cm}
\begin{itemize}
    \vspace{-0.1cm}
    \item We propose a group-based OOD detection framework, along with a novel OOD scoring function MOS, that scales substantially better for large label space. Our method establishes the new state-of-the-art performance, reducing the average FPR95 by \textbf{14.33}\%  while achieving \textbf{6x} speedup in inference time compared to the best baseline.
    \vspace{-0.2cm}
    \item We conduct extensive ablations which improve understandings of our method for
large-scale OOD detection under (1) different grouping strategies, (2) different sizes of semantic class space, (3) different backbone architectures, and (4) varying fine-tuning capacities.
 \vspace{-0.2cm}
    \item We curate diverse OOD evaluation datasets from four real-world high-resolution image databases, which enables future research to evaluate OOD detection methods in a large-scale setting\footnote{ Code and data for reproducing our results are available at:~\url{https://github.com/deeplearning-wisc/large_scale_ood}}.
\end{itemize}






\section{Preliminary and Analysis}
\vspace{-0.1cm}
\paragraph{Preliminaries} We consider a training dataset drawn i.i.d.\ from the in-distribution $P_{\bm{X}}$, with label space ${Y} = \{1,2,\cdots,C \}$. For OOD detection problem, it is common to train a classifier $f(\mathbf{x})$ on the in-distribution $P_{\bm{X}}$, and evaluate on samples that are drawn from a different distribution $Q_{\bm{X}}$. An OOD detector $G(\mathbf{x})$ is a binary classifier:
\vspace{-0.1cm}
\begin{equation*}
    G(\mathbf{x}) = 
    \begin{cases}
    \text{in}, &\text{if}\ S(\mathbf{x}) \geq \gamma \\
    \text{out}, &\text{if}\ S(\mathbf{x}) < \gamma,
    \end{cases}
\end{equation*}
where $S(\mathbf{x})$ is the scoring function, and $\gamma$ is the threshold chosen so that a high fraction (\eg, 95\%) of in-distribution data is correctly classified. 

\begin{figure}[h]
    \centering
    \includegraphics[width=0.5\textwidth]{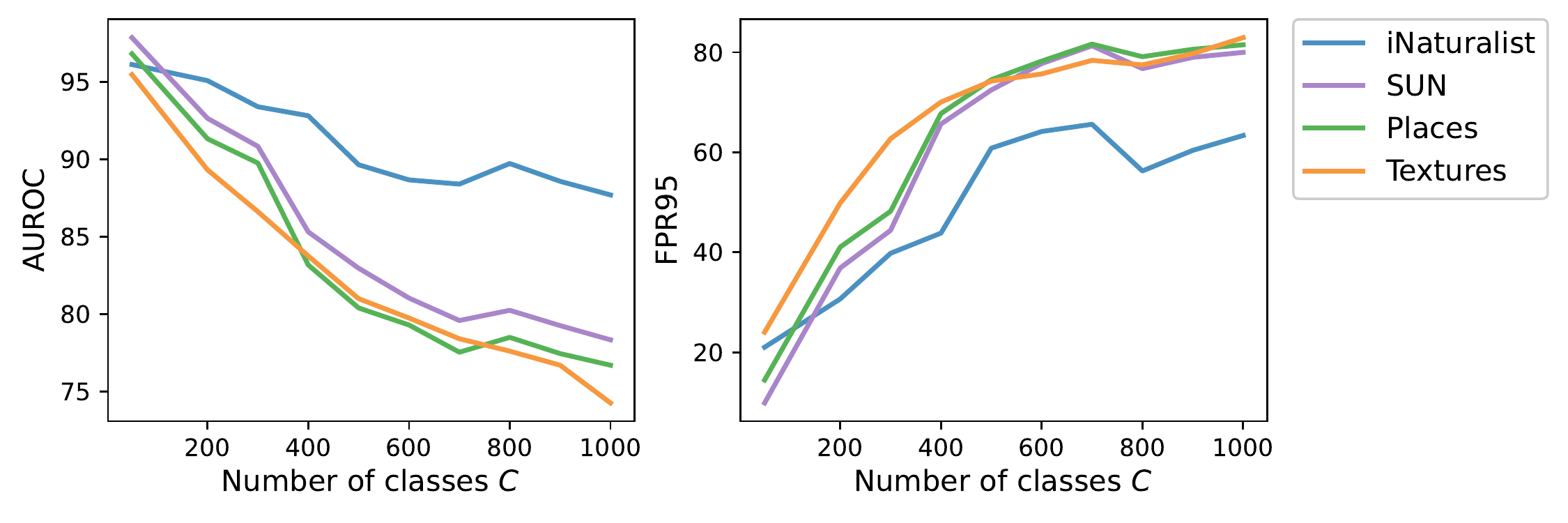}
    \caption{\small OOD detection performance of a common baseline MSP~\cite{hendrycks2016baseline} decreases rapidly as the number of ImageNet-1k classes increases (\emph{left}: AUROC; \emph{right}: FPR95).} 
    \label{fig:class_num_ablation_baseline}
\end{figure}

\vspace{-0.3cm}
\paragraph{Effect of Number of Classes on OOD Detection} We first revisit the common baseline approach~\cite{hendrycks2016baseline}, which uses the maximum softmax probability (MSP), $S(\*x)=\max_i \frac{e^{f_i(\*x)}}{\sum_{j=1}^C e^{f_j(\*x)}}$, for OOD detection. We investigate the effect of label space size on the OOD detection performance. In particular, we use a ResNetv2-101 architecture~\cite{he2016identity} trained on different subsets\footnote{To create the training subset, we first randomly select $C$ ($C \in \{50, 200, 300, 400, 500, 600, 700, 800, 900, 1000\}$) labels from the 1,000 ImageNet classes. For each of the chosen label, we then sample 700 images for training.} of ImageNet with varying numbers of classes $C$. As shown in Figure~\ref{fig:class_num_ablation_baseline}, the performance (FPR95) degrades rapidly from 17.34\% to 76.94\% as the number of in-distribution classes increases from 50 to 1,000. 
This trend signifies that current OOD detection methods are indeed challenged by the
increasingly large label space, which motivates our work. 

\begin{figure}[t]
    \centering
    \includegraphics[width=0.54\textwidth]{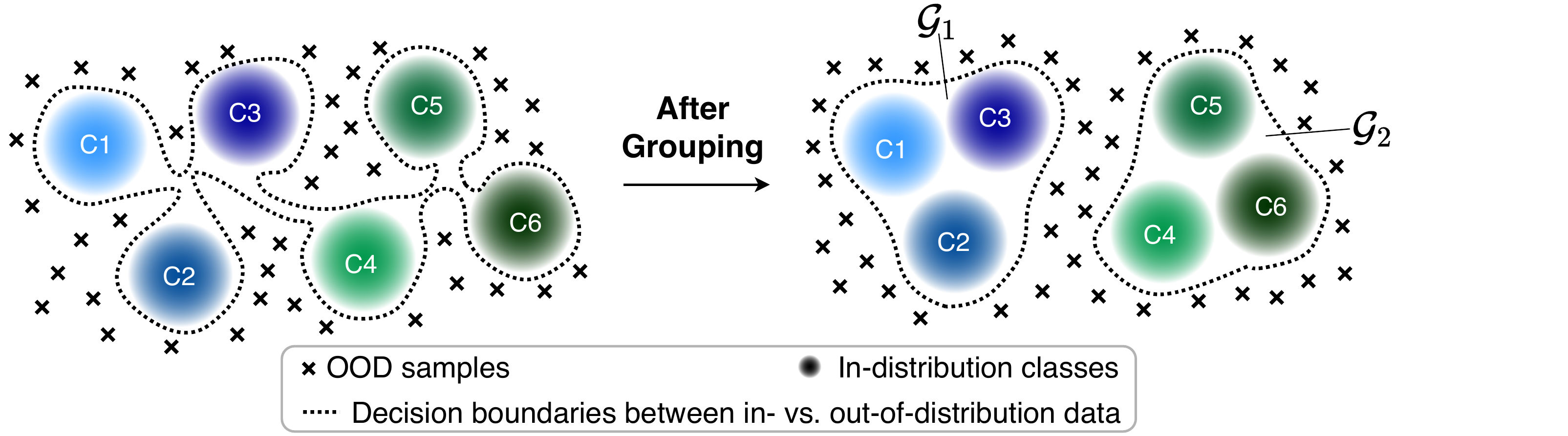}
    \caption{\small{A toy example in 2D space of group-based OOD detection framework. \emph{Left}: without grouping, the decision boundary between in- vs. out-of-distribution data becomes increasingly complex with more classes. \emph{Right}: our group-based method simplifies the decision boundary and reduces the uncertainty space for OOD data.}}
    \vspace{-0.5cm}
    \label{fig:2d_toy_example}
\end{figure}

\vspace{-0.4cm}

\section{Method}

Our novel group-based OOD detection framework is illustrated
in Figure~\ref{fig:main_arch}. In what follows, we first provide an overview and then describe the group softmax training technique in Section~\ref{sec:group-softmax}. We introduce our proposed OOD detection algorithm MOS in Section~\ref{sec:gs-ood}, followed with grouping strategies in Section~\ref{sec:grouping}. 


\vspace{-0.3cm}
\paragraph{Method Overview: A Conceptual Example} As aforementioned, OOD detection performance can suffer notably from the increasing number of in-distribution classes. To mitigate this issue, our key idea is to decompose the large semantic space into smaller groups with similar concepts, which allows simplifying the decision boundary and reducing the uncertainty space between in- vs. out-of-distribution data. We illustrate our idea with a toy example in Figure~\ref{fig:2d_toy_example}, where the in-distribution data consists of  class-conditional Gaussians. Without grouping (left), the decision boundary between in- vs. OOD data is determined by \emph{all} classes and becomes increasingly complex as the number of classes grows. In contrast, with grouping (right), the decision boundary for OOD detection can be significantly simplified, as shown by the dotted curves. 

In other words, by way of grouping, the OOD detector only needs to make a small number of relatively simple estimations about \textit{whether an image belongs to this group}, as opposed to making a large number of hard decisions about \textit{whether an image belongs to this class}.  An image will be classified as OOD if it belongs to none of the groups. We proceed with describing the training mechanism that achieves our novel conceptual idea.  

\subsection{Group-based Learning}
\label{sec:group-softmax}
We divide the total number of $C$ categories into $K$ groups, $\mathcal{G}_1, \mathcal{G}_2, ..., \mathcal{G}_K$. We calculate the standard group-wise softmax for each group $\mathcal{G}_k$:
\vspace{-0.2cm}
\begin{equation}
    p_c^k (\mathbf{x}) = \frac{e^{f_c^k (\mathbf{x})}}{\sum_{c' \in \mathcal{G}_k} e^{f_{c'}^k (\mathbf{x})}}, \ c \in \mathcal{G}_k,
\vspace{-0.1cm}
\end{equation}
where $f_c^k (\mathbf{x})$ and $p_c^k (\mathbf{x})$ denote the output logit and the softmax probability for class $c$ in group $\mathcal{G}_k$, respectively.

\vspace{-0.2cm}
\paragraph{Category ``Others''} Standard group softmax is insufficient as it can only discriminate classes within the group, but cannot estimate the OOD uncertainty between inside vs. outside the group. To this end, a new category \texttt{others} is introduced to every group, as shown in Figure~\ref{fig:main_arch}. The model can predict \texttt{others} if the input $\*x$ does not belong to this group. In other words, the \texttt{others} category allows explicitly learning the decision boundary between inside vs. outside the group, as illustrated by the dashed curves surrounding classes C1/C2/C3 in Figure~\ref{fig:2d_toy_example}. This is desirable for OOD detection, as an OOD input can be mapped to \texttt{others} for all groups, whereas an in-distribution input will be mapped to one of the semantic categories in some group with high confidence.

Importantly, our use of the category \texttt{others}  creates ``virtual" group-level outlier data 
without relying on any external data. Each training example $\*x$ not only helps estimate the decision boundary for the classification problem, but also effectively improves the OOD uncertainty estimation for groups to which it does not belong. We show the formulation can in fact achieve the dual objective of in-distribution classification, as well as OOD detection. 


\vspace{-0.3cm}
\paragraph{Training and Inference} During training, the ground-truth labels are re-mapped in each group. In groups where $c$ is not included, class \texttt{others} will be defined as the ground-truth class. 
The training objective is a sum of cross-entropy losses in each group:
\vspace{-0.2cm}
\begin{equation}
    \mathcal{L}_{GS} = - \frac{1}{N}\sum_{n=1}^N \sum_{k=1}^K \sum_{c\in \mathcal{G}_k}y_c^k\log (p_c^k (\mathbf{x})),
\vspace{-0.1cm}
\end{equation}
where $y_c^k$ and $p_c^k$ represent the label and the softmax probability of category $c$ in $\mathcal{G}_k$, and $N$ is the total number of training samples.

 We denote the set of all valid (non-\texttt{others}) classes in each group as $\mathcal{G}'_k = \mathcal{G}_k \backslash \{\text{\texttt{others}}\}$. During inference time, we derive the group-wise class prediction in the valid set for each group:

\vspace{-0.2cm}
\begin{equation*}
    \hat p^k = \max_{c \in \mathcal{G}'_k} p_c^k(\mathbf{x}),\ \hat c^k = \argmax_{c \in \mathcal{G}'_k} p_c^k(\mathbf{x}).
    \vspace{-0.1cm}
\end{equation*}
Then we use the maximum group-wise softmax score and the corresponding class for final prediction:
\vspace{-0.2cm}
\begin{equation*}
    k_* = \argmax_{1 \leq k \leq K} \hat p^k.
\vspace{-0.1cm}
\end{equation*}
The final prediction is category $\hat c^{k_*}$ from group $\mathcal{G}_{k_*}$.



\subsection{OOD Detection with MOS}
\label{sec:gs-ood}
For a classification model trained with the group softmax loss, we propose a novel OOD scoring function, \textbf{Minimum Others Score (MOS)}, that allows effective differentiation between in- vs. out-of-distribution data. Our key observation is that category \texttt{others} carries useful information for how likely an image is OOD with respect to each group. 

As discussed in Section~\ref{sec:group-softmax}, an OOD input will be mapped to \texttt{others} with high confidence in all groups, whereas an in-distribution input will have a low score on category \texttt{others} in the group it belongs to. Therefore, the lowest \texttt{others} score among all groups is crucial for distinguishing between in- vs. out-of-distribution data. This leads to the following OOD scoring function, termed as \emph{Minimum Others Score}:
\vspace{-0.2cm}
\begin{equation}
    S_\text{MOS}(\mathbf{x}) = -\min_{1 \leq k \leq K}{p_{\text{others}}^k(\mathbf{x})}.
\vspace{-0.1cm}
\end{equation}
Note that we negate the sign to align with the conventional notion that $S_\text{MOS}(\*x)$ is higher for in-distribution data and lower for out-of-distribution.

\begin{figure}[h]
    \centering
    \includegraphics[width=0.5\textwidth]{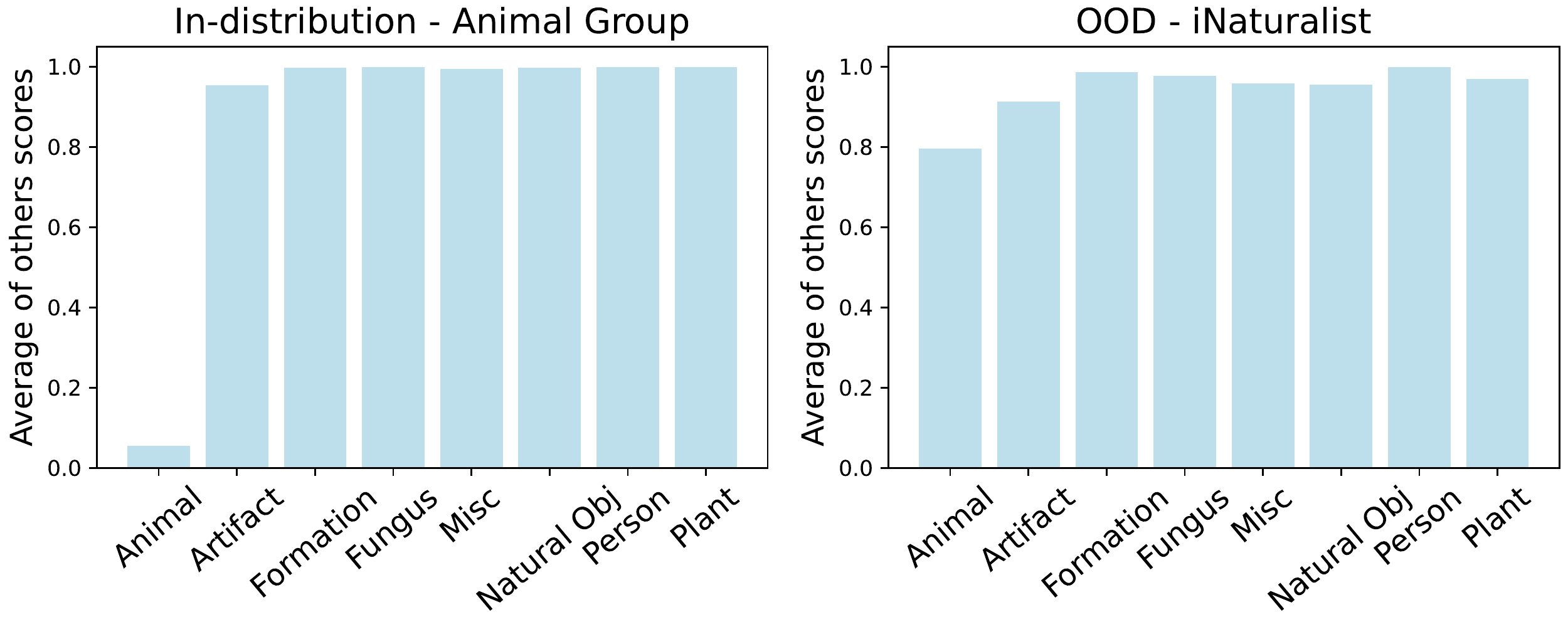}
    \caption{\small{Average of \texttt{others} scores in each group for both in-distribution data (\textit{left}) and OOD data (\textit{right}).}} 
    \label{fig:score_dist_sample}
    \vspace{-0.4cm}
\end{figure}

To provide an interpretation and intuition behind MOS, we show in Figure~\ref{fig:score_dist_sample} the average scores for the category \texttt{others} in each group for both in-distribution and OOD images. For in-distribution, we select all validation images from the \texttt{animal} group in the ImageNet-1k dataset. The minimum \texttt{others} score among all groups is significantly lower for in-distribution data than that for OOD data, allowing for effective differentiation between them.

\begin{table*}[ht]
\footnotesize{
\centering
\begin{tabular}{l|c|cc|cc|cc|cc|cc}
\toprule
\multirow{3}{*}{\textbf{Method}} & \multirow{3}{*}{\textbf{\begin{tabular}[c]{@{}c@{}}Test \\ Time\\ (min)\end{tabular}}} & \multicolumn{2}{c|}{\textbf{iNaturalist}}    & \multicolumn{2}{c|}{\textbf{SUN}}            & \multicolumn{2}{c|}{\textbf{Places}}         & \multicolumn{2}{c|}{\textbf{Textures}}       & \multicolumn{2}{c}{\textbf{Average}}        \\ \cline{3-12} 
                                 &                                                                                        & \textbf{AUROC}       & \textbf{FPR95}        & \textbf{AUROC}       & \textbf{FPR95}        & \textbf{AUROC}       & \textbf{FPR95}        & \textbf{AUROC}       & \textbf{FPR95}        & \textbf{AUROC}       & \textbf{FPR95}       \\
                                 &                                                                                        & \multicolumn{1}{c}{$\uparrow$} & \multicolumn{1}{c|}{$\downarrow$} & \multicolumn{1}{c}{$\uparrow$} & \multicolumn{1}{c|}{$\downarrow$} & \multicolumn{1}{c}{$\uparrow$} & \multicolumn{1}{c|}{$\downarrow$} & \multicolumn{1}{c}{$\uparrow$} & \multicolumn{1}{c|}{$\downarrow$} & \multicolumn{1}{c}{$\uparrow$} & \multicolumn{1}{c}{$\downarrow$}  \\ \midrule
MSP~\cite{hendrycks2016baseline}                              & 3.1                                                                                    & 87.59                & 63.69                 & 78.34                & 79.98                 & 76.76                & 81.44                 & 74.45                & 82.73                 & 79.29                & 76.96                \\
ODIN~\cite{liang2018enhancing}                             & 23.6                                                                                   & 89.36                & 62.69                 & 83.92                & 71.67                 & 80.67                & 76.27                 & 76.30                & 81.31                 & 82.56                & 72.99                \\
Mahalanobis~\cite{lee2018simple}                      & 145.4                                                                                  & 46.33                & 96.34                 & 65.20                & 88.43                 & 64.46                & 89.75                 & 72.10                & 52.23                 & 62.02                & 81.69                \\
Energy~\cite{liu2020energy}                           & 3.1                                                                                    & 88.48                & 64.91                 & 85.32                & 65.33                 & 81.37                & 73.02                 & 75.79                & 80.87                 & 82.74                & 71.03                \\
KL Matching~\cite{hendrycks2019benchmark}                      & 20.6                                                                                   & 93.00                & 27.36                 & 78.72                & 67.52                 & 76.49                & 72.61                 & \textbf{87.07}       & \textbf{49.70}        & 83.82                & 54.30                \\ \midrule
\textbf{MOS (ours)}                       & 3.2                                                                                    & \textbf{98.15}       & \textbf{9.28}         & \textbf{92.01}       & \textbf{40.63}        & \textbf{89.06}       & \textbf{49.54}        & 81.23                & 60.43                 & \textbf{90.11}       & \textbf{39.97}       \\ \bottomrule
\end{tabular}

}
\caption{\small{OOD detection performance comparison between MOS and baselines. All methods are fine-tuned from the same pre-trained BiT-S-R101x1 backbone with ImageNet-1k as in-distribution dataset. The description of 4 OOD test datasets is provided in Section~\ref{sec:dataset}. $\uparrow$ indicates larger values are better, while $\downarrow$ indicates smaller values are better. All values are percentages. \textbf{Bold} numbers are superior results. Test time for all methods are evaluated with the same in- and out-of-distribution datasets (60k images in total). }} 
\label{table:main_result}
\vspace{-0.3cm}
\end{table*}

\subsection{Grouping Strategies}
\label{sec:grouping}

Given the dependency on the group structure, a natural question arises: \emph{how do different grouping strategies affect the performance of OOD detection}? To answer this, we systematically consider three grouping strategies: (1) taxonomy, (2) feature clustering, and (3) random grouping. 

\vspace{-0.4cm}
\paragraph{Taxonomy} The first grouping strategy is applicable when the taxonomy of the label space is known. For example, in the case of ImageNet, each class is associated with a synset in WordNet~\cite{miller1995wordnet}, from which we can build the taxonomy as a hierarchical tree. In particular, we adopt the 8 super-classes defined by ImageNet\footnote{\url{http://image-net.org/explore}} as our groups and map each category into one of the 8 groups: \texttt{animal}, \texttt{artifact}, \texttt{geological formation}, \texttt{fungus}, \texttt{misc}, \texttt{natural object}, \texttt{person}, and \texttt{plant}.

\vspace{-0.4cm}
\paragraph{Feature Clustering} When taxonomy is not available, we can approximately estimate the structure of semantic classes through feature clustering. Specifically, we extract feature representations for each training image from a pre-trained feature extractor. Then, the feature representation of each class is the average of feature embeddings in that class. Finally, we perform a K-Means clustering~\cite{macqueen1967some} on categorical feature representations, one for each class.

\vspace{-0.4cm}
\paragraph{Random Grouping} Lastly, we contrast the taxonomy and the feature clustering strategies with random grouping, where each class is randomly assigned to a group. This allows us to estimate the lower bound of OOD detection performance with MOS.

By default, we use taxonomy as the grouping strategy if not specified otherwise. In Section~\ref{sec:grouping_method_ablation}, we experimentally compare the OOD detection performance using all three grouping strategies.



\section{Experiments}
\vspace{-0.1cm}
We first describe the evaluation datasets (Section~\ref{sec:dataset}) and  experimental setups (Section~\ref{sec:exp_setup}). In Section~\ref{sec:exp_results}, we show that MOS achieves state-of-the-art OOD detection performance, followed by extensive ablations that improve the understandings of MOS for large-scale OOD detection.

\subsection{Datasets}
\label{sec:dataset}

\subsubsection{In-distribution Dataset}
\vspace{-0.2cm}
We use ImageNet-1k~\cite{deng2009imagenet} as the in-distribution dataset, which covers a wide range of real-world objects. ImageNet-1k has at least 10 times more labels compared to CIFAR datasets used in prior literature. In addition, the image resolution is also significantly higher than CIFAR (32$\times$32) and MNIST (28$\times$28).

\vspace{-0.4cm}
\subsubsection{Out-of-distribution Datasets}
\vspace{-0.2cm}
To evaluate our approach, we consider a diverse collection of OOD test datasets, spanning various domains including fine-grained images, scene images, and textural images. We carefully curate the OOD evaluation benchmarks to make sure concepts in these datasets do not overlap with ImageNet-1k. Below we describe the construction of each evaluation dataset in detail. Samples of each OOD dataset are provided in Figure~\ref{fig:main_arch}. We provide the list of concepts chosen for each OOD dataset in Appendix~\ref{app:ood_class}.

\vspace{-0.4cm}
\paragraph{iNaturalist} iNaturalist~\cite{van2018inaturalist} is a fine-grained dataset containing 859,000 images across more than 5,000 species of plants and animals. All images are resized to have a max dimension of 800 pixels. We manually select 110 plant classes not present in ImageNet-1k, and randomly sample 10,000 images for these 110 classes. 
\vspace{-0.4cm}
\paragraph{SUN} SUN~\cite{xiao2010sun} is a scene database of 397 categories and 130,519 images with sizes larger than $200\times200$. 
SUN and ImageNet-1k have overlapping categories. Therefore, we carefully select 50 nature-related concepts that are unique in SUN,  such as \textit{forest} and \textit{iceberg}. We randomly sample 10,000 images for these 50 classes. 
\vspace{-0.4cm}
\paragraph{Places} Places365~\cite{zhou2017places} is another scene dataset with similar concept coverage as SUN. 
All images in this dataset have been resized to have a minimum dimension of 512. We manually select 50 categories from this dataset that are not present in ImageNet-1k and then randomly sample 10,000 images for these 50 categories.

\vspace{-0.5cm}
\paragraph{Textures}  Textures~\cite{cimpoi2014describing} consists of 5,640 images of textural patterns, with sizes ranging between $300\times 300$ and $640 \times 640$. We use the entire dataset for evaluation. 


\begin{figure*}[t]
    \centering
    \vspace{-0.3cm}
    \includegraphics[width=0.99\textwidth]{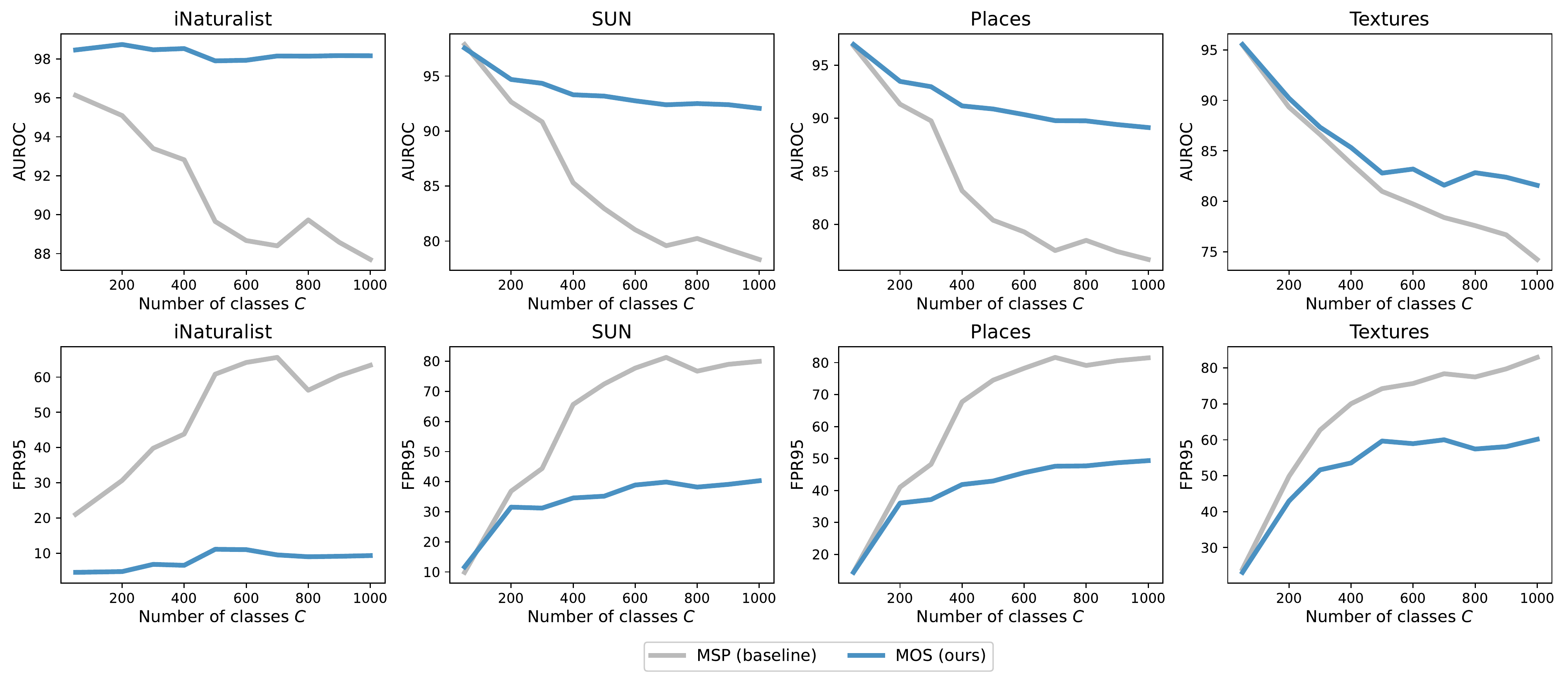}
    \caption{\small{OOD detection performance of MOS (blue) and the MSP baseline (gray). MOS exhibits more stabilized performance as the number of in-distribution classes increases. For each OOD dataset, we show AUROC (\textit{top}) and FPR95 ( \textit{bottom}).}}
    \label{fig:exp_class_num_ablation_fix_each}
    \vspace{-0.3cm}
\end{figure*}

\subsection{Experiment Setup}
\label{sec:exp_setup}
\vspace{-0.1cm}
\paragraph{Pre-trained Backbone} We use Google BiT-S models~\cite{kolesnikov2020big} as our feature extractor in all experiments. The models are trained on ImageNet-1k, with ResNetv2 architectures~\cite{he2016identity} at varying capacities.
Pre-trained models allow extracting high-quality features with minimal time and energy consumption. In practice, one can always choose to train from scratch.

For the main results, we use the BiT-S-R101x1 model with depth 101 and width factor 1, unless specified otherwise. We provide a comparison of using  feature extractors of varying model sizes in Section~\ref{sec:extractor_ablation}. For efficiency, we fix the backbone and only fine-tune the last fully-connected (FC) layer in the main experiments. We additionally explore the effect of fine-tuning more layers beyond the last FC layer in Section~\ref{sec:finetune_ablation}.

\vspace{-0.5cm}
\paragraph{Training Details} We follow the procedure in BiT-HyperRule~\cite{kolesnikov2020big} and fine-tune the pre-trained BiT-S model for 20k steps with a batch size of 512. We use SGD with an initial learning rate of 0.003 and a momentum of 0.9. The learning rate is decayed by a factor of 10 at 30\%, 60\%, and 90\% of the training steps. During training, all images are resized to 512 $\times$ 512 and randomly cropped to 480 $\times$ 480. At test time, all images are resized to 480 $\times$ 480. A learning rate warm-up is used for the first 500 steps. We perform all experiments on NVIDIA GeForce RTX 2080Ti GPUs.

\vspace{-0.5cm}
\paragraph{Evaluation Metrics} We measure the following metrics that are commonly used for OOD detection: (1)~the false positive rate of OOD examples when the true positive rate of in-distribution examples is at 95\% (FPR95); (2)~the area under the receiver operating characteristic curve (AUROC). We additionally report the area under the precision-recall curve (AUPR) in Appendix~\ref{app:aupr}.

\vspace{-0.1cm}
\subsection{Results}
\label{sec:exp_results}

\subsubsection{MOS vs. Existing Methods}
\label{sec:exp_main_results}
\vspace{-0.2cm}
The main results are shown in Table~\ref{table:main_result}. We report performance for each dataset described in Section~\ref{sec:dataset}, as well as the average performance. For fair evaluation, we compare with competitive methods in the literature that derive OOD scoring functions from a model trained on in-distribution data and do not rely on auxiliary outlier data. We first compare with approaches driven by small datasets, including MSP~\cite{hendrycks2016baseline}, ODIN~\cite{liang2018enhancing}, Mahalanobis~\cite{lee2018simple}, as well as Energy~\cite{liu2020energy}. All these methods rely on networks trained with flat softmax. Under the same network backbone (BiT-S-R101x1), MOS outperforms the best baseline Energy~\cite{liu2020energy} by \textbf{31.06}\% in FPR95. It is also worth noting that fine-tuning with group softmax maintains competitive classification accuracy (75.16\%) on in-distribution data compared with its flat softmax counterpart (75.20\%).

We also compare our method with KL matching~\cite{hendrycks2019benchmark}, a competitive baseline evaluated on large-scale image classification. MOS reduces FPR95 by \textbf{14.33}\% compared to KL matching. Note that for each input, KL matching needs to calculate its KL divergence to all class centers.
Therefore, the running time of KL matching increases linearly with the number of in-distribution categories, which can be computationally expensive for a very large label space. As shown in Table~\ref{table:main_result}, our method achieves a \textbf{6x} speedup compared to KL matching.

\vspace{-0.3cm}
\subsubsection{MOS with Increasing Numbers of Classes}
\label{sec:class_num_ablation}
\vspace{-0.2cm}
In Figure~\ref{fig:exp_class_num_ablation_fix_each}, we show the OOD detection performance as we increase the number of in-distribution classes $C\in \{50, 200, 300, 400, 500, 600, 700, 800, 900, 1000\}$ on ImageNet-1k. For each $C$, we create training data by first randomly sampling $C$ labels from the entire 1k classes, and then sampling 700 images for each chosen label.  
Importantly, we observe that MOS (in blue) exhibits more stabilized performance as $C$ increases, compared to MSP~\cite{hendrycks2016baseline} (in gray). For example, on the iNaturalist OOD dataset, FPR95 rises from 21.02\% to 63.36\% using MSP, whilst MOS degrades by only 4.76\%. This trend signifies that MOS is an effective approach for scaling OOD detection towards a large semantic space. 

We also explore an alternative setting where we fix the total number of training images, as we vary the number of classes $C$. In this setting, the model is trained on fewer images per class as the number of classes increases, making the problem even more challenging. We  report those results in Appendix~\ref{app:class_num_ablation_fix_total}. Overall, MOS remains less sensitive to the number of classes compared to the MSP baseline.


\begin{figure}[t]
    \centering
    \vspace{-0.3cm}
    \includegraphics[width=0.45\textwidth]{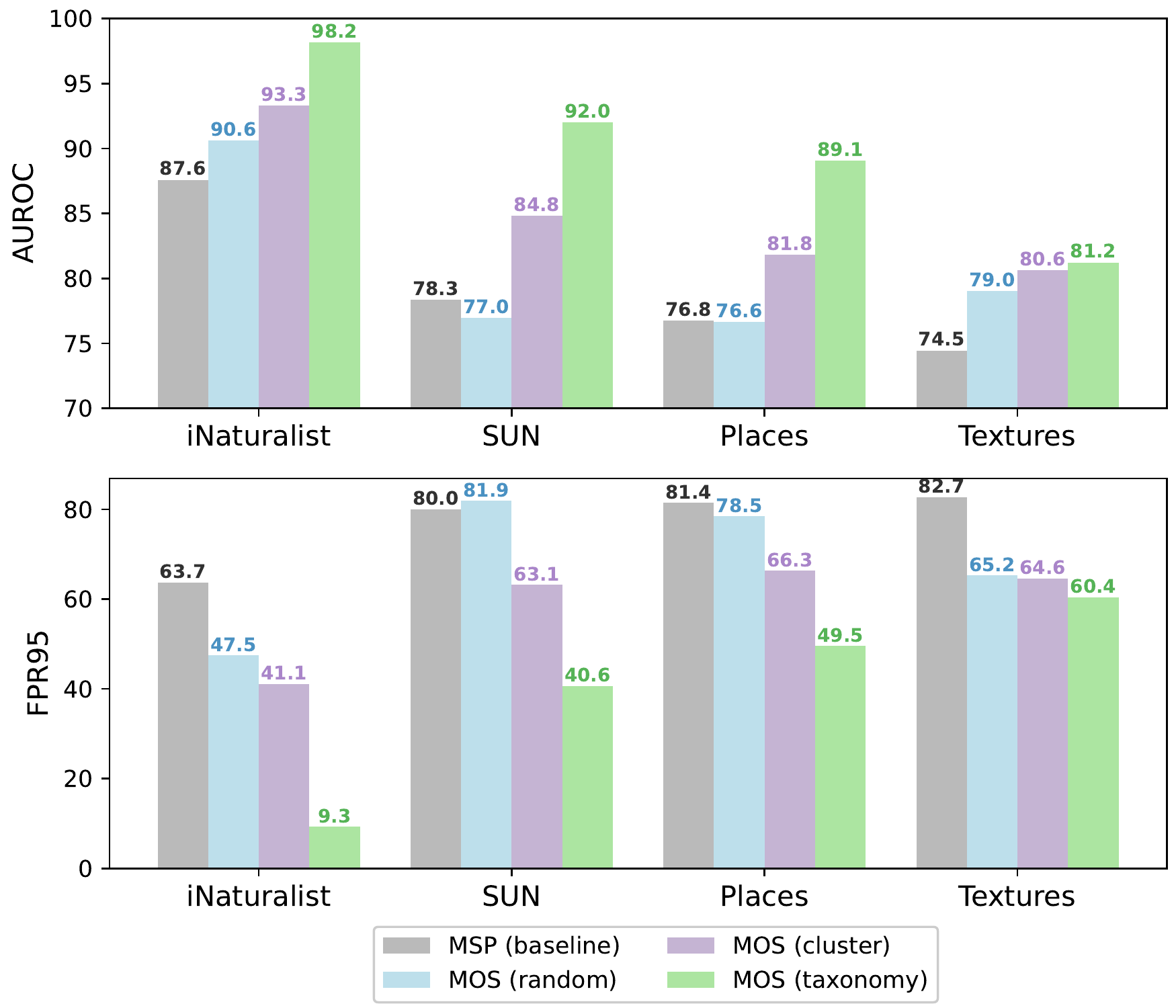}
    \caption{\small{OOD detection performance comparison between MOS with different grouping strategies and the MSP baseline on 4 OOD datasets. (\textit{top}: AUROC; \textit{bottom}: FPR95).}}
    \label{fig:exp_group_method_ablation}
    \vspace{-0.3cm}
\end{figure}

\vspace{-0.3cm}
\subsubsection{MOS with Different Grouping Strategies}
\vspace{-0.2cm}
\label{sec:grouping_method_ablation}
In this ablation, we contrast the performance of three different grouping strategies described in Section~\ref{sec:grouping}. For a fair comparison, we use the number of groups $K=8$ for all methods, since ImageNet taxonomy has 8 super-classes.
For feature clustering, we first extract the feature vector from the penultimate layer of the pre-trained BiT-S model for each training image. The feature representation for each category is the average feature vector among all images in that category. We then perform a K-Means clustering on the 1,000 categorical feature vectors (one for each class) with $K=8$. For random grouping, we randomly split 1,000 classes into 8 groups with equal sizes (125 classes each). 

We compare the performance of MOS under different grouping strategies in Figure~\ref{fig:exp_group_method_ablation}. We observe that feature clustering works substantially better than the MSP baseline~\cite{hendrycks2016baseline} while maintaining similar in-distribution classification accuracy (-0.16\%) to the taxonomy-based grouping. Interestingly, random grouping achieves better performance than the MSP baseline~\cite{hendrycks2016baseline} on 3 out of 4 OOD datasets. However, we do observe a drop of in-distribution classification accuracy (-0.98\%) using random grouping, compared to the taxonomy-based grouping. We argue that feature clustering is a viable strategy when taxonomy is unavailable, as it outperforms MSP by 18.2\% (FPR95) on average. We additionally report how different numbers of groups $K$ affect the OOD detection performance for all three grouping strategies in Appendix~\ref{app:group_num_ablation}. 

\begin{figure*}[t]
    \centering
        \vspace{-0.2cm}
    \includegraphics[width=0.98\textwidth]{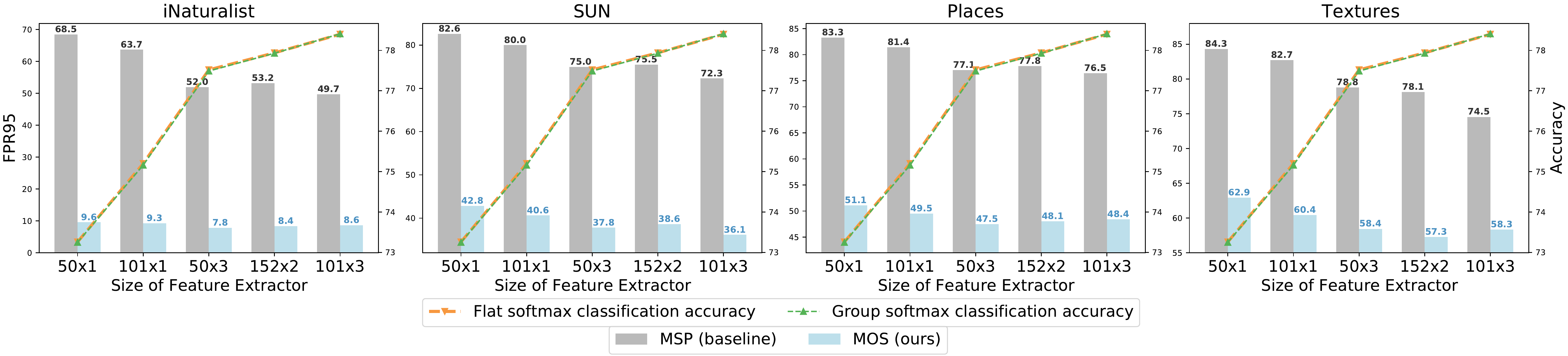}
    \caption{\small{Effect of using different pre-trained feature extractors. The x-axis indicates feature extractors with larger capacities from left to right. Only the top FC layer is fine-tuned in all experiments. Both OOD detection (\textit{bars}) and image classification (\textit{dashed lines}) benefit from improved feature extractors.}}
    \vspace{-0.2cm}
    \label{fig:exp_feature_extractor_ablation}
\end{figure*}

\subsubsection{MOS with Different Feature Extractors}
\label{sec:extractor_ablation}
\vspace{-0.2cm}
We investigate how the performance of OOD detection changes as we employ different pre-trained feature extractors.
In Figure~\ref{fig:exp_feature_extractor_ablation},  we compare the performance of using a family of 5 feature extractors (in increasing size): BiT-S-R50x1, BiT-S-R101x1, BiT-S-R50x3, BiT-S-R152x2, BiT-S-R101x3\footnote{\url{https://github.com/google-research/big_transfer}}. All models are ResNetv2 architectures with varying depths and width factors. It is important to note that since we fix the entire backbone and only fine-tune the last FC layer, this ablation is about the effect of the quality of feature extractors rather than model capacities.

As we use feature extractors trained on larger capacities, the classification accuracy increases, with comparable performance between using the flat vs. group softmax. Overall the OOD detection performance improves as the capacity of feature extractors increases. More importantly, MOS consistently outperforms MSP~\cite{hendrycks2016baseline} in all cases. 
These results suggest that using pre-trained models with better feature representations will not only improve classification accuracy but also benefit OOD detection performance.

\begin{figure*}[t]
    \centering
    \includegraphics[width=0.98\textwidth]{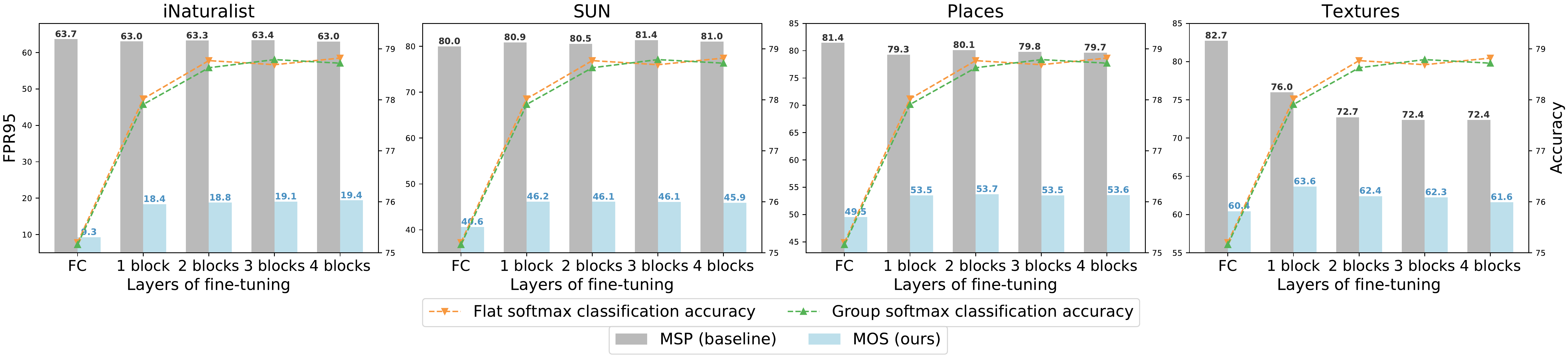}
    \caption{\small{Effect of fine-tuning different numbers of residual blocks in BiT-S-R101x1. We show both OOD detection (\textit{bars}) and image classification (\textit{dashed lines}) performance.}}
    \vspace{-0.2cm}
    \label{fig:exp_finetune_layer_ablation}
\end{figure*}

\vspace{-0.4cm}
\subsubsection{MOS with Varying Fine-tuning Capacities}
\label{sec:finetune_ablation}
\vspace{-0.2cm}
In this ablation, we explore the efficacy of fine-tuning more layers. Concretely, we go beyond the FC layer and fine-tune different numbers of residual blocks in BiT-S-R101x1. Figure~\ref{fig:exp_finetune_layer_ablation} shows the classification accuracy and OOD detection performance under different fine-tuning capacities. Noticeably, MOS consistently outperforms MSP~\cite{hendrycks2016baseline} in OOD detection under all fine-tuning capacities. 
As expected, we observe that fine-tuning more layers leads to better classification accuracy. However, increasing the number of fine-tuning layers would adversely affect OOD detection in some cases.
We hypothesize that fine-tuning with more layers will result in more label-overfitted predictions, and undesirably produce higher confidence scores for OOD data. This suggests that only fine-tuning the top FC layers is not only computationally efficient but also in fact desirable for OOD detection performance. 


\section{Related Work}

\paragraph{OOD Detection with Pre-trained Models} Hendrycks and Gimpel~\cite{hendrycks2016baseline} establish a common baseline for OOD detection by using maximum softmax probability (MSP). Several works attempt to improve the OOD uncertainty estimation by using ODIN score~\cite{liang2018enhancing}, deep ensembles~\cite{lakshminarayanan2017simple}, Mahalanobis distance-based confidence score~\cite{lee2018simple}, generalized ODIN score~\cite{hsu2020generalized}, and energy score~\cite{liu2020energy}. Lin et al.~\cite{lin2021mood} propose a dynamic OOD inference framework that improved the computational efficiency. However, previous methods driven by small datasets are  suboptimal  in a large-scale setting. In contrast, MOS scales better with large label space, outperforming existing methods by a large margin.

\vspace{-0.4cm}
\paragraph{OOD Detection with Model Fine-tuning} An orthogonal line of work explores training with auxiliary outlier data for model regularization~\cite{bevandic2018discriminative,geifman2019selectivenet,malinin2018predictive,mohseni2020self,subramanya2017confidence, liu2020energy}. Auxiliary outlier data can either be realistic images~\cite{hendrycks2018deep,mohseni2020self,papadopoulos2019outlier,liu2020energy,chen2020robust-new} or synthetic images generated by GANs~\cite{lee2018training}. Several loss functions have been designed to regularize model predictions of the auxiliary outlier data towards uniform distributions~\cite{lee2018training}, a background class for OOD data~\cite{chen2020robust-new, mohseni2020self}, or higher energies~\cite{liu2020energy}. 
In this work, our model is fine-tuned only on in-distribution data, as we do not assume the availability of auxiliary outlier data. Different from previous settings, it can be prohibitive to construct an auxiliary outlier dataset in large-scale image classification, since the in-distribution data has a much wider coverage of concepts.

\paragraph{Generative Modeling Based OOD Detection} Generative models~\cite{kingma2013auto,tabak2013family,rezende2014stochastic,dinh2017density,van2016conditional} estimate the probability density of input data and can thus be directly utilized as OOD detectors with high density indicating in-distribution and low density indicating out-of-distribution. However, as shown in ~\cite{nalisnick2018deep}, deep generative models can undesirably assign a high likelihood to OOD data. Several strategies have been proposed to mitigate this issue, such as improved metrics~\cite{choi2018generative}, likelihood ratio~\cite{ren2019likelihood,serra2019input}, and modified training techniques~\cite{hendrycks2018deep}. In this work, we mainly focus on the discriminative-based approaches. It is important to note that generative models~\cite{hinz2018generating} can be prohibitively challenging to train and optimize on large-scale real-world datasets.

\vspace{-0.4cm}
\paragraph{OOD Detection for Large-scale Classification}
Several works make pioneering efforts in large-scale OOD detection. Roady~\etal~\cite{roady2019outofdistribution} sample half of the classes from ImageNet-1k as in-distribution data, and evaluate the other half as OOD test data. They use a one-vs-rest training strategy and background class regularization, which requires access to an auxiliary dataset. KL matching was employed as the OOD scoring function in~\cite{hendrycks2019benchmark}. In this work, we propose a novel group-based solution that scales more effectively and efficiently for large-scale OOD detection. We also perform evaluations on more diverse real-world OOD datasets and conduct thorough ablations that improve understandings of the problem and solutions in many aspects.

\vspace{-0.4cm}
\paragraph{Learning with Hierarchical Labels} The hierarchical structure of the class categories has been utilized for efficient inference~\cite{deng2011fast,liu2013probabilistic}, improved classification accuracy~\cite{deng2014large}, and stronger object detection performance~\cite{redmon2017yolo9000}. Some works aim to learn a label tree structure when taxonomy is unavailable~\cite{bengio2010label,deng2011fast,liu2013probabilistic}. As a typical hierarchy, group-based learning has been widely adopted in image classification tasks~\cite{hinton2015distilling,ahmed2016network,yan2015hd,warde2014self,gross2017hard}. Recently, a group softmax classifier is proposed to tackle the problem of long-tail object detection, where categories are grouped according to the number of training instances~\cite{li2020overcoming}. We contribute to this field by showing the promise of using a group label structure for effective OOD detection.




\vspace{-0.2cm}
\section{Conclusion}
\vspace{-0.1cm}
In this paper, we propose a group-based OOD detection framework, along with a novel OOD scoring function, {MOS}, that effectively scales the OOD detection to a real-world setting with a large label space. 
We curate four diverse OOD evaluation datasets that allow future research to evaluate OOD detection methods in a large-scale setting. Extensive experiments show our group-based framework can significantly improve the performance of OOD detection in this large-scale setting compared to existing approaches. We hope our research can raise more attention to expand the view of OOD detection from small benchmarks to a large-scale real-world setting.


{\small

\bibliographystyle{ieee_fullname}
\bibliography{main}
}

\newpage
\appendix
\onecolumn


\begin{center}
      {\Large \bf {MOS: Towards Scaling Out-of-distribution Detection for Large Semantic Space
      
      (Supplementary Material)} \par}
     
      \vskip .5em
      \vspace*{12pt}
\end{center}

\section{Selected Categories in OOD Datasets}
\label{app:ood_class}
To evaluate our approach, we consider a diverse collection of OOD test datasets, spanning various domains. We carefully curate the OOD evaluation benchmarks to make sure  concepts  in  these  datasets do not overlap with ImageNet-1k~\cite{deng2009imagenet}. Below we provide the list of concepts chosen for each OOD dataset, including iNaturalist~\cite{van2018inaturalist}, SUN~\cite{xiao2010sun}, Places365~\cite{zhou2017places}, and Textures~\cite{cimpoi2014describing}. We hope the information would allow future research to reproduce our results.

\vspace{-0.4cm}
\paragraph{iNaturalist} \textit{Coprosma lucida}, \textit{Cucurbita foetidissima}, \textit{Mitella diphylla}, \textit{Selaginella bigelovii}, \textit{Toxicodendron vernix}, \textit{Rumex obtusifolius}, \textit{Ceratophyllum demersum}, \textit{Streptopus amplexifolius}, \textit{Portulaca oleracea}, \textit{Cynodon dactylon}, \textit{Agave lechuguilla}, \textit{Pennantia corymbosa}, \textit{Sapindus saponaria}, \textit{Prunus serotina}, \textit{Chondracanthus exasperatus}, \textit{Sambucus racemosa}, \textit{Polypodium vulgare}, \textit{Rhus integrifolia}, \textit{Woodwardia areolata}, \textit{Epifagus virginiana}, \textit{Rubus idaeus}, \textit{Croton setiger}, \textit{Mammillaria dioica}, \textit{Opuntia littoralis}, \textit{Cercis canadensis}, \textit{Psidium guajava}, \textit{Asclepias exaltata}, \textit{Linaria purpurea}, \textit{Ferocactus wislizeni}, \textit{Briza minor}, \textit{Arbutus menziesii}, \textit{Corylus americana}, \textit{Pleopeltis polypodioides}, \textit{Myoporum laetum}, \textit{Persea americana}, \textit{Avena fatua}, \textit{Blechnum discolor}, \textit{Physocarpus capitatus}, \textit{Ungnadia speciosa}, \textit{Cercocarpus betuloides}, \textit{Arisaema dracontium}, \textit{Juniperus californica}, \textit{Euphorbia prostrata}, \textit{Leptopteris hymenophylloides}, \textit{Arum italicum}, \textit{Raphanus sativus}, \textit{Myrsine australis}, \textit{Lupinus stiversii}, \textit{Pinus echinata}, \textit{Geum macrophyllum}, \textit{Ripogonum scandens}, \textit{Echinocereus triglochidiatus}, \textit{Cupressus macrocarpa}, \textit{Ulmus crassifolia}, \textit{Phormium tenax}, \textit{Aptenia cordifolia}, \textit{Osmunda claytoniana}, \textit{Datura wrightii}, \textit{Solanum rostratum}, \textit{Viola adunca}, \textit{Toxicodendron diversilobum}, \textit{Viola sororia}, \textit{Uropappus lindleyi}, \textit{Veronica chamaedrys}, \textit{Adenocaulon bicolor}, \textit{Clintonia uniflora}, \textit{Cirsium scariosum}, \textit{Arum maculatum}, \textit{Taraxacum officinale officinale}, \textit{Orthilia secunda}, \textit{Eryngium yuccifolium}, \textit{Diodia virginiana}, \textit{Cuscuta gronovii}, \textit{Sisyrinchium montanum}, \textit{Lotus corniculatus}, \textit{Lamium purpureum}, \textit{Ranunculus repens}, \textit{Hirschfeldia incana}, \textit{Phlox divaricata laphamii}, \textit{Lilium martagon}, \textit{Clarkia purpurea}, \textit{Hibiscus moscheutos}, \textit{Polanisia dodecandra}, \textit{Fallugia paradoxa}, \textit{Oenothera rosea}, \textit{Proboscidea louisianica}, \textit{Packera glabella}, \textit{Impatiens parviflora}, \textit{Glaucium flavum}, \textit{Cirsium andersonii}, \textit{Heliopsis helianthoides}, \textit{Hesperis matronalis}, \textit{Callirhoe pedata}, \textit{Crocosmia $\times$ crocosmiiflora}, \textit{Calochortus albus}, \textit{Nuttallanthus canadensis}, \textit{Argemone albiflora}, \textit{Eriogonum fasciculatum}, \textit{Pyrrhopappus pauciflorus}, \textit{Zantedeschia aethiopica}, \textit{Melilotus officinalis}, \textit{Peritoma arborea}, \textit{Sisyrinchium bellum}, \textit{Lobelia siphilitica}, \textit{Sorghastrum nutans}, \textit{Typha domingensis}, \textit{Rubus laciniatus}, \textit{Dichelostemma congestum}, \textit{Chimaphila maculata}, \textit{Echinocactus texensis}

\vspace{-0.4cm}

\paragraph{SUN} \textit{badlands}, \textit{bamboo forest}, \textit{bayou}, \textit{botanical garden}, \textit{canal (natural)}, \textit{canal (urban)}, \textit{catacomb}, \textit{cavern (indoor)}, \textit{corn field}, \textit{creek}, \textit{crevasse}, \textit{desert (sand)}, \textit{desert (vegetation)}, \textit{field (cultivated)}, \textit{field (wild)}, \textit{fishpond}, \textit{forest (broadleaf)}, \textit{forest (needleleaf)}, \textit{forest path}, \textit{forest road}, \textit{hayfield}, \textit{ice floe}, \textit{ice shelf}, \textit{iceberg}, \textit{islet}, \textit{marsh}, \textit{ocean}, \textit{orchard}, \textit{pond}, \textit{rainforest}, \textit{rice paddy}, \textit{river}, \textit{rock arch}, \textit{sky}, \textit{snowfield}, \textit{swamp}, \textit{tree farm}, \textit{trench}, \textit{vineyard}, \textit{waterfall (block)}, \textit{waterfall (fan)}, \textit{waterfall (plunge)}, \textit{wave}, \textit{wheat field}, \textit{herb garden}, \textit{putting green}, \textit{ski slope}, \textit{topiary garden}, \textit{vegetable garden}, \textit{formal garden}

\vspace{-0.4cm}

\paragraph{Places} \textit{badlands}, \textit{bamboo forest}, \textit{canal (natural)}, \textit{canal (urban)}, \textit{corn field}, \textit{creek}, \textit{crevasse}, \textit{desert (sand)}, \textit{desert (vegetation)}, \textit{desert road}, \textit{field (cultivated)}, \textit{field (wild)}, \textit{field road}, \textit{forest (broadleaf)}, \textit{forest path}, \textit{forest road}, \textit{formal garden}, \textit{glacier}, \textit{grotto}, \textit{hayfield}, \textit{ice floe}, \textit{ice shelf}, \textit{iceberg}, \textit{igloo}, \textit{islet}, \textit{japanese garden}, \textit{lagoon}, \textit{lawn}, \textit{marsh}, \textit{ocean}, \textit{orchard}, \textit{pond}, \textit{rainforest}, \textit{rice paddy}, \textit{river}, \textit{rock arch}, \textit{ski slope}, \textit{sky}, \textit{snowfield}, \textit{swamp}, \textit{swimming hole}, \textit{topiary garden}, \textit{tree farm}, \textit{trench}, \textit{tundra}, \textit{underwater (ocean deep)}, \textit{vegetable garden}, \textit{waterfall}, \textit{wave}, \textit{wheat field}

\vspace{-0.4cm}
\paragraph{Textures} all images in this dataset
\section{More Ablation Studies}
\subsection{MOS with Increasing Numbers of Classes (A More Challenging Setting)}
\label{app:class_num_ablation_fix_total}
\begin{figure*}[t]
    \centering
    \includegraphics[width=0.99\textwidth]{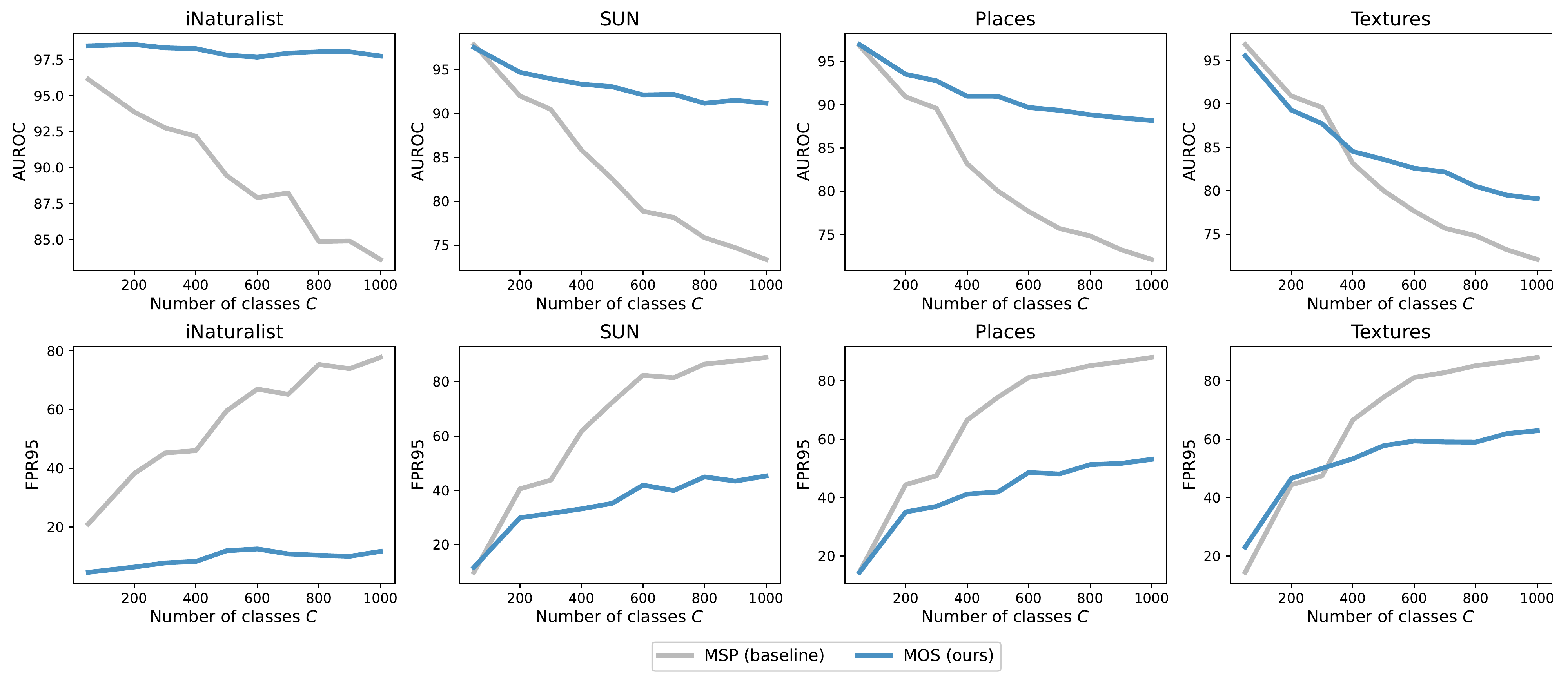}
    \caption{\small{OOD detection performance of MOS (blue) and the MSP baseline (gray). MOS exhibits more stabilized performance as the number of in-distribution classes increases. For each OOD dataset, we show AUROC (\textit{top}) and FPR95 ( \textit{bottom}). Different from Figure~\ref{fig:exp_class_num_ablation_fix_each}, we fix the number of \emph{total training images} instead of the number of \emph{training images in each category} in this experiment.}}
    \vspace{-0.1cm}
    \label{fig:exp_class_num_ablation_fix_total}
\end{figure*}

In Section~\ref{sec:class_num_ablation}, we increase the number of in-distribution classes while fixing the number of \emph{training images in each class} and observe the degradation of OOD detection performance. Here we investigate an alternative setting where we fix the number of \emph{total training images} to be $35,000$, as we increase the number of classes $C\in \{50, 200, 300, 400, 500, 600, 700, 800, 900, 1000\}$. For each $C$, we create training data by first randomly sampling $C$ labels from the entire 1,000 classes in ImageNet-1k, and then sampling 35,000 / $C$ images for each chosen label. In Figure~\ref{fig:exp_class_num_ablation_fix_total}, we show the OOD detection performance with varying numbers of in-distribution classes $C$.

\begin{figure*}[t]
    \centering
    \includegraphics[width=0.88\textwidth]{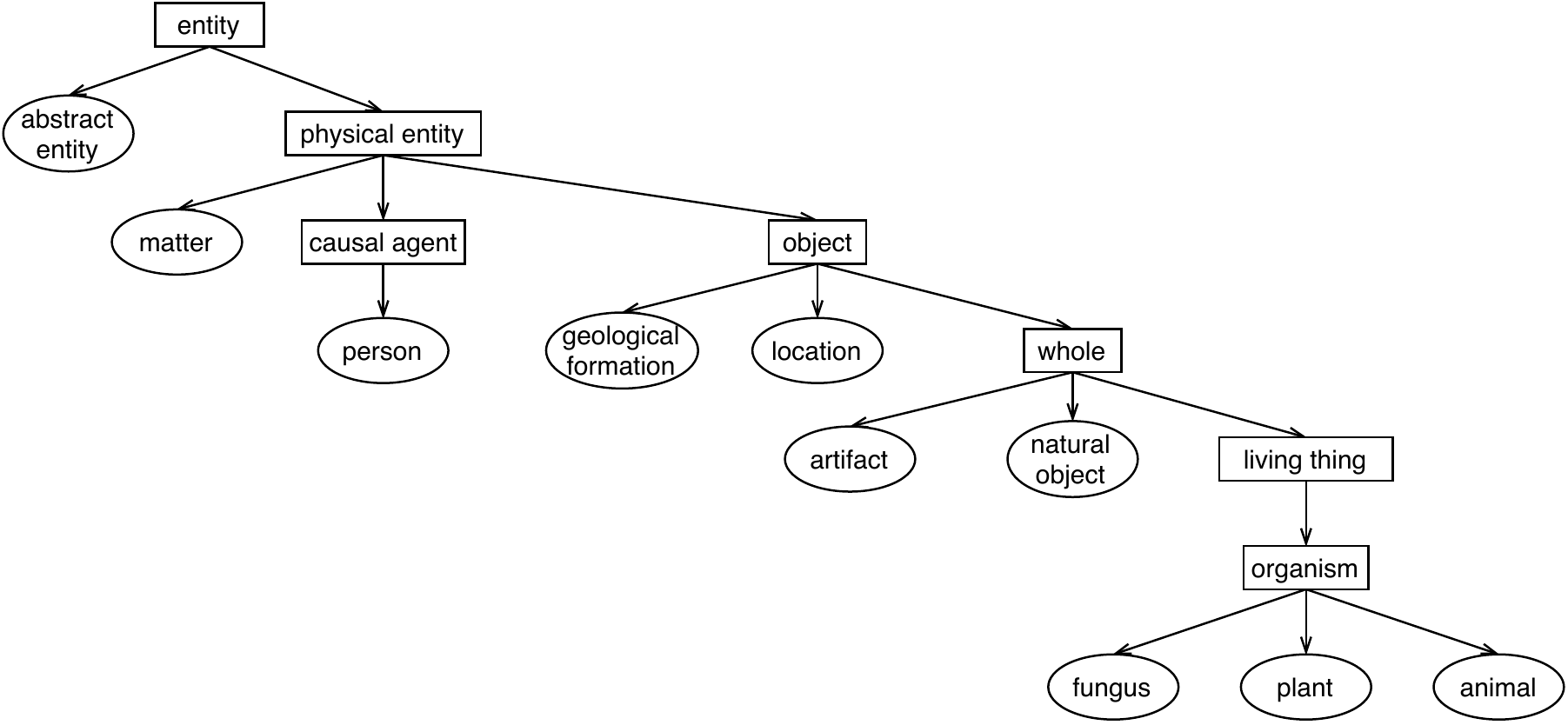}
    \caption{\small{WordNet hierarchy. Super-classes of ImageNet-1k are based on the leaf nodes (in ellipses) except for \texttt{misc}. The super-class of \texttt{misc} contains 3 leaf nodes: \texttt{abstract entity}, \texttt{matter}, and \texttt{location}.}}
    \label{fig:superclass_tree}
    \vspace{-0.5cm}
\end{figure*}

This is a more challenging setting because when the number of classes increases, there will be fewer training images for each class. Unsurprisingly, the performance degradation of OOD detection is more severe in this setting. For instance, on the iNaturalist OOD dataset, the FPR95 performance of MSP~\cite{hendrycks2016baseline} degrades by 56.74\% when the number of 
classes increases from 50 to 1,000, while the corresponding degradation is only 42.34\% in the previous setting. Importantly, MOS remains much less sensitive to the change of the number of in-distribution classes compared to the MSP baseline (without grouping). In particular, on the Places OOD dataset, the FPR95 performance drops from 14.45\% to 88.02\% using MSP, while MOS degrades by only 38.73\%. 

\subsection{MOS with Varying Numbers of Groups}
\label{app:group_num_ablation}

\begin{table*}[t]
\centering
{\scriptsize
\begin{tabular}{c|c|l|cc|cc|cc|cc|cc}
\toprule
\multirow{3}{*}{\textbf{Level}} & \multirow{3}{*}{\textbf{\begin{tabular}[c]{@{}c@{}}Number\\ of\\ Groups\end{tabular}}} & \multicolumn{1}{c|}{\multirow{3}{*}{\textbf{\begin{tabular}[c]{@{}c@{}}Grouping \\ Strategy\end{tabular}}}} & \multicolumn{2}{c|}{\textbf{iNaturalist}}    & \multicolumn{2}{c|}{\textbf{SUN}}            & \multicolumn{2}{c|}{\textbf{Places}}         & \multicolumn{2}{c|}{\textbf{Textures}}       & \multicolumn{2}{c}{\textbf{Average}}        \\ \cline{4-13} 
                                &                                                                                        & \multicolumn{1}{c|}{}                                                                                       & \textbf{AUROC}       & \textbf{FPR95}        & \textbf{AUROC}       & \textbf{FPR95}        & \textbf{AUROC}       & \textbf{FPR95}        & \textbf{AUROC}       & \textbf{FPR95}        & \textbf{AUROC}       & \textbf{FPR95}       \\
                                &                                                                                        & \multicolumn{1}{c|}{}                                                                                       & \multicolumn{1}{c}{$\uparrow$} & \multicolumn{1}{c|}{$\downarrow$} & \multicolumn{1}{c}{$\uparrow$} & \multicolumn{1}{c|}{$\downarrow$} & \multicolumn{1}{c}{$\uparrow$} & \multicolumn{1}{c|}{$\downarrow$} & \multicolumn{1}{c}{$\uparrow$} & \multicolumn{1}{c|}{$\downarrow$} & \multicolumn{1}{c}{$\uparrow$} & \multicolumn{1}{c}{$\downarrow$}  \\ \midrule
\multirow{3}{*}{-3}             & \multirow{3}{*}{2}                                                                     & taxonomy                                                                                                    & \textbf{97.66}       & \textbf{11.20}        & 85.27                & 65.86                 & 82.21                & 70.31                 & 79.06                & 63.88                 & 86.05                & \textbf{52.81}       \\
                                &                                                                                        & feature clustering                                                                                          & 94.91                & 33.12                 & \textbf{87.87}       & \textbf{57.02}        & \textbf{84.31}       & \textbf{66.23}        & \textbf{84.58}       & \textbf{56.06}        & \textbf{87.92}       & 53.11                \\
                                &                                                                                        & random grouping                                                                                             & 91.12                & 46.82                 & 79.28                & 78.73                 & 78.64                & 76.50                 & 79.02                & 63.53                 & 82.02                & 66.40                \\ \midrule
\multirow{3}{*}{-2}             & \multirow{3}{*}{3}                                                                     & taxonomy                                                                                                    & \textbf{97.21}       & \textbf{15.78}        & \textbf{92.28}       & \textbf{40.08}        & \textbf{89.35}       & \textbf{49.74}        & 81.00                & 60.64                 & \textbf{89.96}       & \textbf{41.56}       \\
                                &                                                                                        & feature clustering                                                                                          & 94.57                & 33.58                 & 87.23                & 57.18                 & 83.60                & 65.34                 & \textbf{83.06}       & \textbf{57.23}        & 87.12                & 53.33                \\
                                &                                                                                        & random grouping                                                                                             & 90.75                & 47.55                 & 76.57                & 83.00                 & 75.89                & 81.33                 & 80.40                & 61.93                 & 80.90                & 68.45                \\ \midrule
\multirow{3}{*}{-1}             & \multirow{3}{*}{6}                                                                     & taxonomy                                                                                                    & \textbf{97.16}       & \textbf{16.16}        & \textbf{92.07}       & \textbf{40.28}        & \textbf{89.12}       & \textbf{49.53}        & 81.34                & \textbf{60.27}        & \textbf{89.92}       & \textbf{41.56}       \\
                                &                                                                                        & feature clustering                                                                                          & 90.68                & 53.49                 & 82.95                & 72.24                 & 79.48                & 75.06                 & \textbf{81.78}       & 62.50                 & 83.72                & 65.82                \\
                                &                                                                                        & random grouping                                                                                             & 91.55                & 44.73                 & 79.11                & 78.99                 & 78.31                & 76.17                 & 80.93                & 62.30                 & 82.48                & 65.55                \\ \midrule
\multirow{3}{*}{0}              & \multirow{3}{*}{8}                                                                     & taxonomy                                                                                                    & \textbf{98.15}       & \textbf{9.28}         & \textbf{92.01}       & \textbf{40.63}        & \textbf{89.06}       & \textbf{49.54}        & \textbf{81.23}       & \textbf{60.43}        & \textbf{90.11}       & \textbf{39.97}       \\
                                &                                                                                        & feature clustering                                                                                          & 93.29                & 41.13                 & 84.84                & 63.15                 & 81.84                & 66.33                 & 80.62                & 64.61                 & 85.15                & 58.81                \\
                                &                                                                                        & random grouping                                                                                             & 90.63                & 47.47                 & 76.95                & 81.89                 & 76.65                & 78.47                 & 79.02                & 65.25                 & 80.81                & 68.27                \\ \midrule
\multirow{3}{*}{1}              & \multirow{3}{*}{36}                                                                    & taxonomy                                                                                                    & \textbf{95.85}       & \textbf{23.73}        & \textbf{90.51}       & \textbf{47.53}        & \textbf{87.74}       & \textbf{52.51}        & \textbf{88.47}       & \textbf{45.55}        & \textbf{90.64}       & \textbf{42.33}       \\
                                &                                                                                        & feature clustering                                                                                          & 91.22                & 47.73                 & 79.25                & 79.48                 & 76.53                & 79.00                 & 82.81                & 61.72                 & 82.45                & 66.98                \\
                                &                                                                                        & random grouping                                                                                             & 91.01                & 46.96                 & 79.66                & 77.79                 & 79.36                & 74.35                 & 78.91                & 68.72                 & 82.24                & 66.96                \\ \midrule
\multirow{3}{*}{2}              & \multirow{3}{*}{85}                                                                    & taxonomy                                                                                                    & 92.22                & 46.19                 & \textbf{88.07}       & \textbf{59.41}        & \textbf{86.02}       & \textbf{60.28}        & \textbf{85.40}       & \textbf{57.70}        & \textbf{87.93}       & \textbf{55.90}       \\
                                &                                                                                        & feature clustering                                                                                          & \textbf{93.13}       & \textbf{40.07}        & 81.05                & 77.06                 & 78.33                & 76.44                 & 82.28                & 64.57                 & 83.70                & 64.54                \\
                                &                                                                                        & random grouping                                                                                             & 90.58                & 50.04                 & 78.75                & 81.21                 & 78.81                & 77.23                 & 76.95                & 76.17                 & 81.27                & 71.16                \\ \midrule
\multirow{3}{*}{3}              & \multirow{3}{*}{225}                                                                   & taxonomy                                                                                                    & 90.35                & 57.38                 & \textbf{85.19}       & \textbf{71.72}        & \textbf{83.57}       & \textbf{69.99}        & \textbf{81.40}       & \textbf{72.27}        & \textbf{85.13}       & \textbf{67.84}       \\
                                &                                                                                        & feature clustering                                                                                          & \textbf{91.49}       & \textbf{48.90}        & 79.59                & 82.16                 & 78.06                & 79.97                 & 79.40                & 75.09                 & 82.14                & 71.53                \\
                                &                                                                                        & random grouping                                                                                             & 89.66                & 56.81                 & 77.55                & 84.73                 & 78.16                & 79.61                 & 75.07                & 82.43                 & 80.11                & 75.90                \\ \midrule
\multirow{3}{*}{4}              & \multirow{3}{*}{416}                                                                   & taxonomy                                                                                                    & 89.18                & 63.48                 & \textbf{82.34}       & 80.60                 & \textbf{81.30}       & \textbf{76.88}        & \textbf{78.37}       & 81.17                 & \textbf{82.80}       & 75.53                \\
                                &                                                                                        & feature clustering                                                                                          & \textbf{91.66}       & \textbf{47.91}        & 80.40                & \textbf{79.40}        & 79.12                & 77.26                 & 78.24                & \textbf{80.00}        & 82.36                & \textbf{71.14}       \\
                                &                                                                                        & random grouping                                                                                             & 88.68                & 61.29                 & 76.94                & 86.01                 & 77.67                & 81.41                 & 73.24                & 86.91                 & 79.13                & 78.91                \\ \midrule
\multirow{3}{*}{5}              & \multirow{3}{*}{642}                                                                   & taxonomy                                                                                                    & 88.10                & 67.11                 & \textbf{80.07}       & 84.04                 & \textbf{79.65}       & 79.89                 & 75.17                & 87.22                 & 80.75                & 79.57                \\
                                &                                                                                        & feature clustering                                                                                          & \textbf{90.45}       & \textbf{55.74}        & 79.83                & \textbf{82.89}        & 79.22                & \textbf{79.18}        & \textbf{75.77}       & \textbf{86.01}        & \textbf{81.32}       & \textbf{75.96}       \\
                                &                                                                                        & random grouping                                                                                             & 88.31                & 63.92                 & 77.09                & 85.94                 & 77.60                & 81.69                 & 72.16                & 89.08                 & 78.79                & 80.16                \\ \midrule
\multirow{3}{*}{6}              & \multirow{3}{*}{789}                                                                   & taxonomy                                                                                                    & 87.39                & 69.61                 & 78.57                & 85.73                 & \textbf{78.64}       & 81.04                 & 73.57                & 89.29                 & 79.54                & 81.42                \\
                                &                                                                                        & feature clustering                                                                                          & \textbf{89.81}       & \textbf{59.68}        & \textbf{78.78}       & \textbf{84.75}        & 78.31                & \textbf{80.82}        & \textbf{74.66}       & \textbf{88.65}        & \textbf{80.39}       & \textbf{78.48}       \\
                                &                                                                                        & random grouping                                                                                             & 88.07                & 65.12                 & 76.91                & 86.65                 & 77.52                & 82.15                 & 71.84                & 89.84                 & 78.59                & 80.94                \\ \bottomrule
\end{tabular}
}
\caption{\small{Effect of different numbers of groups on OOD detection performance for 3 grouping strategies (taxonomy, feature clustering, and random grouping). Level 0 represents the level of super-classes in the taxonomy tree (main setting). Positive levels indicate splitting the super-classes into more groups (tracing down the taxonomy tree), while negative levels indicate merging the super-classes into fewer groups (tracing up the taxonomy tree).}}
\label{table:group_number_ablation}
\vspace{-0.4cm}
\end{table*}

In this ablation we investigate how different numbers of groups $K$  affect the OOD detection performance of MOS under three grouping strategies: (1) taxonomy, (2) feature clustering, and (3) random grouping. For taxonomy-based grouping, in order to increase the number of groups, we split the nodes of each super-class into their descendants in the label tree and map the 1,000 classes into one of the descendants instead of the super-classes themselves; in order to decrease the number of groups, we merge some of the super-classes into one group based on Figure~\ref{fig:superclass_tree}. Specifically, we construct 10 taxonomy levels with increasing numbers of groups based on the label tree in the following way:

\vspace{-0.4cm}
\paragraph{Level -3} There are 2 groups in Level -3: \{\texttt{animal}, \texttt{plant}, \texttt{fungus}\}, \{\texttt{artifact}, \texttt{natural object}, \texttt{geological formation}, \texttt{person}, \texttt{misc}\}.

\vspace{-0.4cm}
\paragraph{Level -2} There are 3 groups in Level -2: \{\texttt{animal}, \texttt{plant}, \texttt{fungus}\}, \{\texttt{artifact}, \texttt{natural object}\}, \{\texttt{geological formation}, \texttt{person}, \texttt{misc}\}.

\vspace{-0.4cm}
\paragraph{Level -1} There are 6 groups in Level -1: \{\texttt{animal}, \texttt{plant}, \texttt{fungus}\}, \{\texttt{artifact}\}, \{\texttt{natural object}\}, \{\texttt{geological formation}\}, \{\texttt{person}\}, \{\texttt{misc}\}.

\vspace{-0.4cm}
\paragraph{Level 0} This is the level of 8 super-classes (main setting).

\vspace{-0.4cm}
\paragraph{Level 1$\sim$6} Groups in Level $i$ are direct children of the nodes in Level $(i-1)$.







\vspace{0.2cm}
For feature clustering and random grouping, we set the numbers of groups to be equal to the corresponding numbers at each of the taxonomy levels for fair comparisons.

As shown in Table~\ref{table:group_number_ablation}, for taxonomy-based grouping, the performance of OOD detection is almost optimal when the number of groups is 8 (Level 0), and further increasing or decreasing the number of groups will not lead to improved performance. Moreover, taxonomy-based grouping outperforms feature clustering and random grouping when $K$ is small and mildly large. However, feature clustering surpasses taxonomy-based grouping when the number of groups is sufficiently large. We hypothesize that as we trace down the label tree, the numbers of categories in each group become more imbalanced, which could adversely impact the performance of OOD detection using taxonomy-based grouping.

\section{AUROC Curves}
\label{app:auroc}

\begin{figure*}[h]
    \centering
    \includegraphics[width=1.0\textwidth]{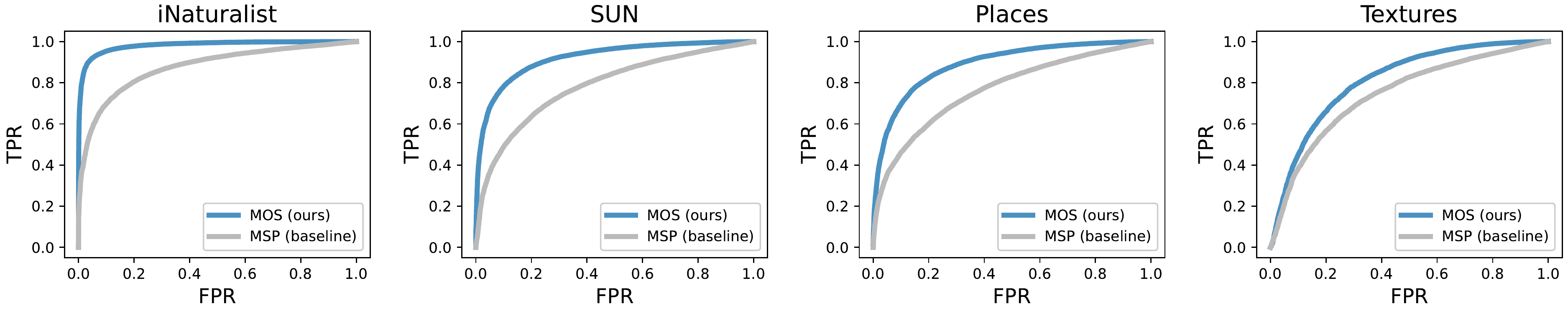}
    \caption{\small{AUROC curves of MOS (blue) and MSP (gray) on four OOD datasets.}}
    \label{fig:auroc}
\end{figure*}

Figure~\ref{fig:auroc} shows the AUROC curves of MOS and MSP for OOD detection. All settings and training details are the same as in Table~\ref{table:main_result}. The gray curve corresponds to the MSP baseline~\cite{hendrycks2016baseline}, while the blue curve corresponds to MOS with taxonomy-based grouping. We observe huge gaps between the gray and the blue AUROC curves on all OOD datasets. For instance, when TPR = 95\%, the FPR can be reduced from 63.69\% to 9.28\% on the iNaturalist OOD dataset.

\section{AUPR Results}
\label{app:aupr}

\begin{table*}[h]
    \centering
    \footnotesize
\begin{tabular}{l|l|ccc}
\toprule
\multirow{2}{*}{\textbf{OOD Dataset}} & \multicolumn{1}{c|}{\multirow{2}{*}{\textbf{Method}}} & \textbf{AUROC}       & \textbf{FPR95}       & \textbf{AUPR}        \\
                                      & \multicolumn{1}{c|}{}                                 & \multicolumn{1}{c}{$\uparrow$} & \multicolumn{1}{c}{$\downarrow$} & \multicolumn{1}{c}{$\uparrow$} \\ \midrule
\multirow{6}{*}{\textbf{iNaturalist}} & MSP                                                   & 87.59                & 63.69                & 97.23                \\
                                      & ODIN                                                  & 89.36                & 62.69                & 97.76                \\
                                      & Mahalanobis                                           & 46.33                & 96.34                & 81.14                \\
                                      & Energy                                                & 88.48                & 64.91                & 97.58                \\
                                      & KL Matching                                           & 93.00                & 27.36                & 97.93                \\
                                      & MOS (ours)                                            & \textbf{98.15}       & \textbf{9.28}        & \textbf{99.62}       \\ \midrule
\multirow{6}{*}{\textbf{SUN}}         & MSP                                                   & 78.34                & 79.98                & 94.45                \\
                                      & ODIN                                                  & 83.92                & 71.67                & 96.26                \\
                                      & Mahalanobis                                           & 65.20                & 88.43                & 88.81                \\
                                      & Energy                                                & 85.32                & 65.33                & 96.57                \\
                                      & KL Matching                                           & 78.72                & 67.52                & 94.10                \\
                                      & MOS (ours)                                            & \textbf{92.01}       & \textbf{40.63}       & \textbf{98.17}       \\ \midrule
\multirow{6}{*}{\textbf{Places}}      & MSP                                                   & 76.76                & 81.44                & 94.15                \\
                                      & ODIN                                                  & 80.67                & 76.27                & 95.35                \\
                                      & Mahalanobis                                           & 64.46                & 89.75                & 88.85                \\
                                      & Energy                                                & 81.37                & 73.02                & 95.49                \\
                                      & KL Matching                                           & 76.49                & 72.61                & 93.61                \\
                                      & MOS (ours)                                            & \textbf{89.06}       & \textbf{49.54}       & \textbf{97.36}       \\ \midrule
\multirow{6}{*}{\textbf{Textures}}    & MSP                                                   & 74.45                & 82.73                & 95.65                \\
                                      & ODIN                                                  & 76.30                & 81.31                & 96.12                \\
                                      & Mahalanobis                                           & 72.10                & 52.23                & 91.89                \\
                                      & Energy                                                & 75.79                & 80.87                & 96.05                \\
                                      & KL Matching                                           & \textbf{87.07}       & \textbf{49.70}       & \textbf{97.97}       \\
                                      & MOS (ours)                                            & 81.23                & 60.43                & 96.65                \\ \midrule
\multirow{6}{*}{\textbf{Average}}     & MSP                                                   & 79.29                & 76.96                & 95.37                \\
                                      & ODIN                                                  & 82.56                & 72.99                & 96.37                \\
                                      & Mahalanobis                                           & 62.02                & 81.69                & 87.67                \\
                                      & Energy                                                & 82.74                & 71.03                & 96.42                \\
                                      & KL Matching                                           & 83.82                & 54.30                & 95.90                \\
                                      & MOS (ours)                                            & \textbf{90.11}       & \textbf{39.97}       & \textbf{97.95}       \\ \bottomrule
\end{tabular}
    \caption{\small{Main results with AUPR. Experimental setups are the same as in Table~\ref{table:main_result}.}}
    \label{tab:main_aupr}
\end{table*}

In Table~\ref{tab:main_aupr} we report the area under the precision-recall curve (AUPR) complementing the AUROC and FPR95 results in Table~\ref{table:main_result}. AUPR is an informative metric in the presence of class imbalance, which is common in OOD detection. Again, MOS demonstrates state-of-the-art performance in terms of AUPR.

\section{\texttt{Others} Scores for All In-distribution Groups and OOD Datasets}
\label{app:all_score_dist}

\begin{figure*}[h]
    \centering
    \includegraphics[width=1.0\textwidth]{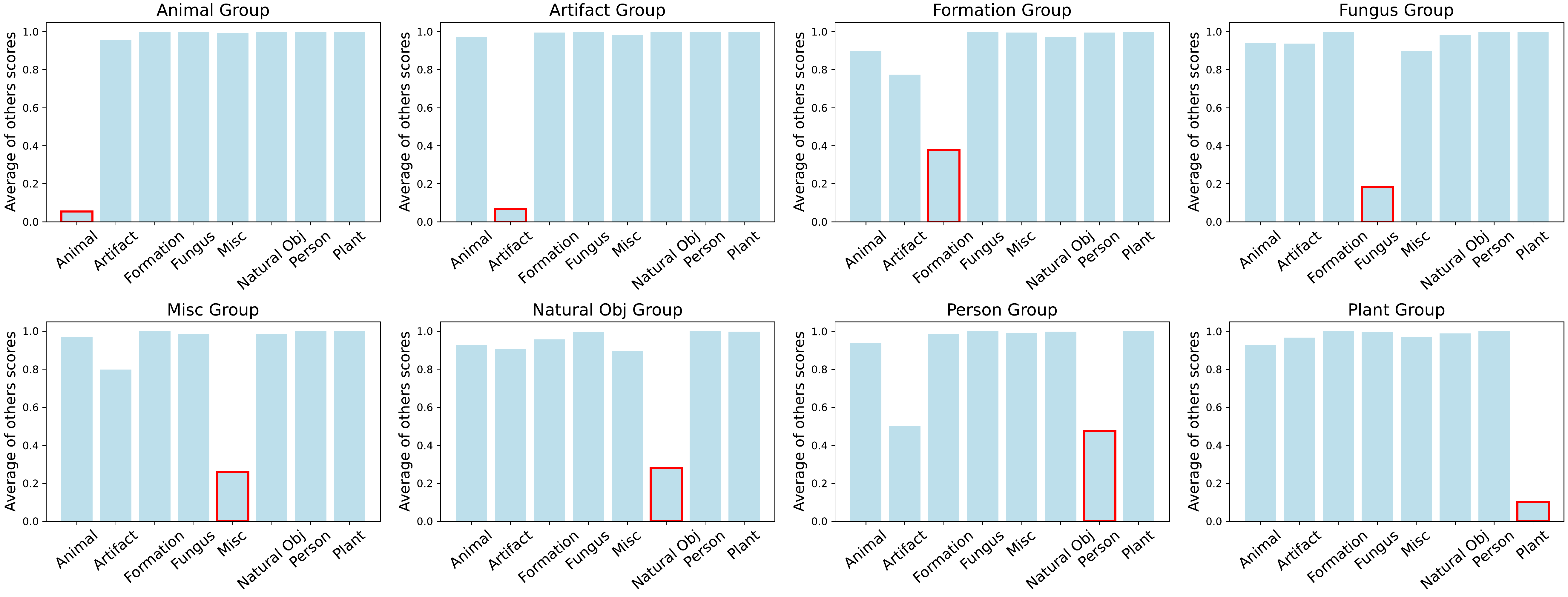}
    \caption{\small{Average \texttt{others} scores for all in-distribution groups. Red boxes indicate the corresponding groups these images belong to.}}
    \label{fig:score_in_dist_groups}
\end{figure*}

\begin{figure*}[h]
    \centering
    \includegraphics[width=1.0\textwidth]{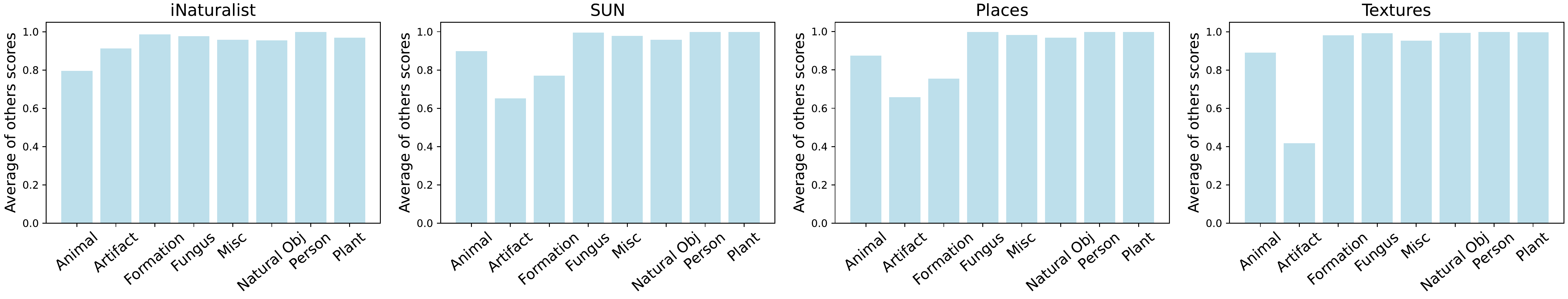}
    \caption{\small{Average \texttt{others} scores for all OOD datasets}}
    \label{fig:score_ood_datasets}
\end{figure*}

Figure~\ref{fig:score_in_dist_groups} and Figure~\ref{fig:score_ood_datasets} show average \texttt{others} scores for 8 in-distribution groups and 4 OOD datasets, respectively. For in-distribution groups, \texttt{others} scores are averaged among all validation images in each group in ImageNet-1k; for OOD datasets, \texttt{others} scores are averaged among all sampled images in the curated datasets.

These histograms provide visual justifications for our method MOS: in-distribution images will have low \texttt{others} scores in at least one group (shown in red boxes), while out-of-distribution images will have high \texttt{others} scores in all 8 groups. Therefore, MOS is effective in distinguishing between in- vs. out-of-distribution data.

\end{document}